\documentclass[review]{elsarticle}
\usepackage{amsmath}
\usepackage{amssymb}
\usepackage{cleveref}
\usepackage{graphicx}
\usepackage{textcomp}
\usepackage{adjustbox}
\usepackage{xcolor}
\usepackage{booktabs}
\usepackage{multirow}

\usepackage[inkscapelatex=false]{svg}
\usepackage{pifont} % for \checkmark
\newcommand{\xmark}{\ding{55}} % for x mark

\usepackage[linesnumbered,ruled,vlined]{algorithm2e}

\journal{}

\biboptions{authoryear}

\begin{document}

\begin{frontmatter}

\title{SynthSeg-Agents: Multi-Agent Synthetic Data Generation for
Zero-Shot Weakly Supervised Semantic Segmentation}
		
		\author[firstaddress,secondaddress]{Wangyu Wu}\ead{v11dryad@foxmail.com}
            \author[thirdaddress]{Zhenhong Chen}\ead{zcheh@microsoft.com}

            \author[secondaddress]{Xiaowei Huang}\ead{xiaowei.huang@liverpool.ac.uk}
		\author[firstaddress]{Fei Ma\corref{mycorrespondingauthor}}\ead{fei.ma@xjtlu.edu.cn}

            \author[firstaddress]{Jimin Xiao\corref{mycorrespondingauthor}}
		\cortext[mycorrespondingauthor]{Corresponding authors} \ead{jimin.xiao@xjtlu.edu.cn}
		
		\address[firstaddress]{Xi'an Jiaotong-Liverpool University, Suzhou, China}
		\address[secondaddress]{University of Liverpool, Liverpool, UK}
        \address[thirdaddress]{Microsoft, Redmond, USA}

\begin{abstract}
Weakly Supervised Semantic Segmentation (WSSS) with image-level labels aims to produce pixel-level predictions without requiring dense annotations. 
While recent approaches have leveraged generative models to augment existing data, they remain dependent on real-world training samples. 
In this paper, we introduce a novel direction—Zero-Shot Weakly Supervised Semantic Segmentation (ZSWSSS)—and propose \textit{SynthSeg-Agents}, a multi-agent framework driven by Large Language Models (LLMs) to generate synthetic training data entirely without real images. 
SynthSeg-Agents comprises two key modules: a \textit{Self-Refine Prompt Agent} and an \textit{Image Generation Agent}. 
The Self-Refine Prompt Agent autonomously crafts diverse and semantically rich image prompts via iterative refinement, memory mechanisms, and prompt-space exploration, guided by CLIP-based similarity and nearest-neighbor diversity filtering.  
These prompts are then passed to the Image Generation Agent, which leverages Vision-Language Models (VLMs) to synthesize candidate images. 
A frozen CLIP scoring model is employed to select high-quality samples, and a ViT-based classifier is further trained to relabel the entire synthetic dataset with improved semantic precision. 
Our framework produces high-quality training data without any real image supervision. 
Experiments on PASCAL VOC 2012 and COCO 2014 show that SynthSeg-Agents achieves competitive performance without using real training images.
This highlights the potential of LLM-driven agents in enabling cost-efficient, scalable semantic segmentation. The code will be made publicly available upon acceptance.
\end{abstract}


\end{frontmatter}

%\linenumbers

\section{Introduction}\label{sec:intro}
Semantic segmentation, as a fundamental task in computer vision, aims to achieve precise pixel-level delineation of target objects. With the rapid advancement of deep learning, although fully supervised segmentation models have demonstrated remarkable success~\cite{chen2017deeplab,yuan2020object,xie2021segformer}, their heavy reliance on costly, densely annotated data significantly constrains model scalability. To mitigate this annotation bottleneck, Weakly Supervised Semantic Segmentation (WSSS) has emerged as a promising alternative that trains segmentation models using weaker forms of supervision, such as image-level labels.
 
\begin{figure}[t]
\centering
\includegraphics[width=0.9\linewidth]{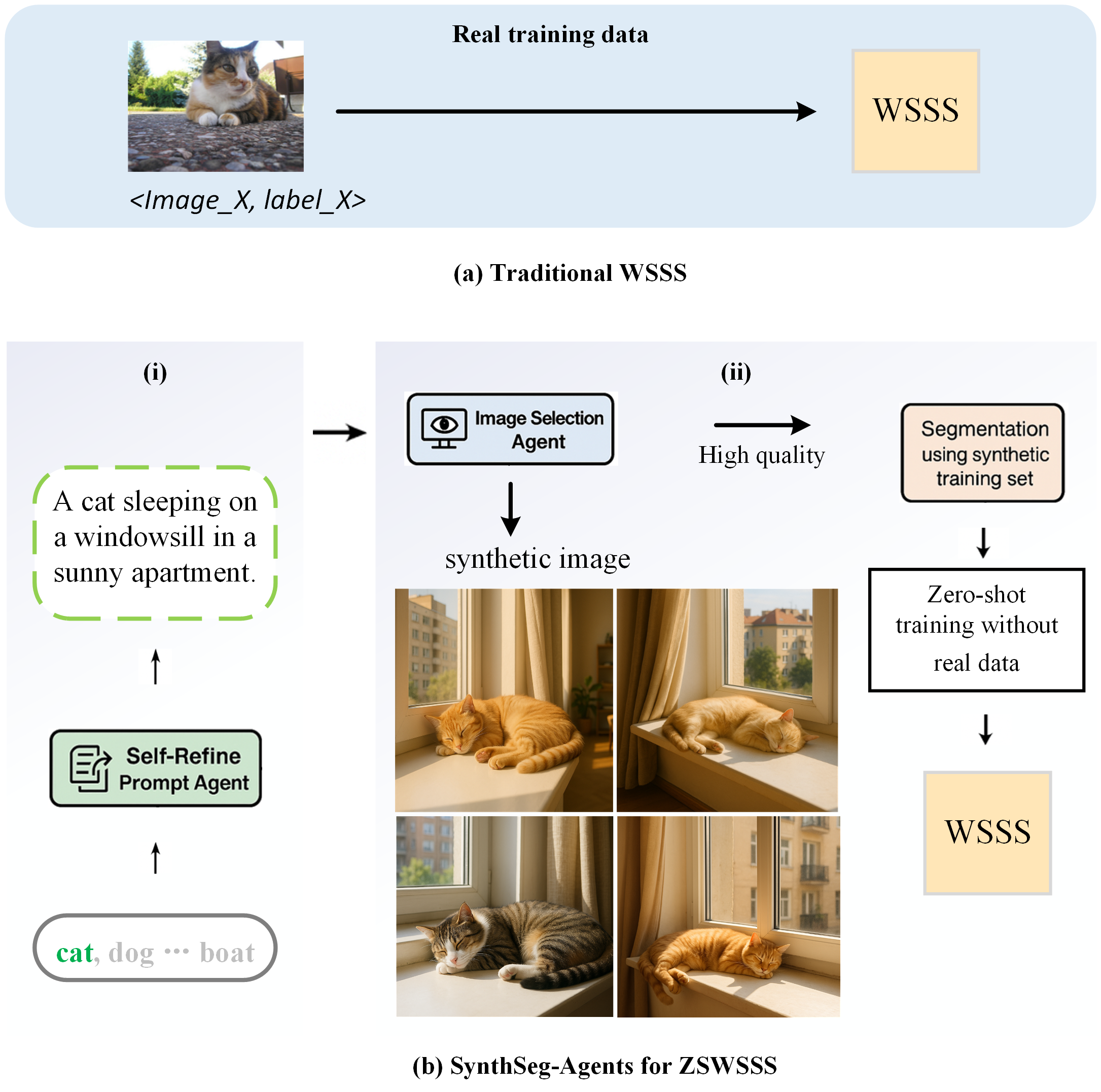}
% \vspace{-0.3cm} 
\caption{Comparison of WSSS frameworks. Traditional methods rely on real images and CAMs, while our SynthSeg-Agents framework generates a synthetic dataset via LLM/VLM collaboration, requiring no real data.}

\label{fig:idea}
\vspace{-0.15cm} 
\end{figure}
Recent WSSS methods have achieved encouraging progress by leveraging Class Activation Maps (CAMs), contrastive learning, and more recently, generative models to augment limited supervision. However, these methods still depend heavily on real training images, which constrains their applicability in scenarios where data collection is difficult, sensitive, or costly. Moreover, the quality of pseudo-labels in WSSS remains bounded by the information content of the available real data. This raises a natural question: \textit{Can we train segmentation models without using any real images at all?}

To address this challenge, we introduce \textit{SynthSeg-Agents} in Fig.~\ref{fig:idea}, a novel multi-agent framework that enables Zero-Shot Weakly Supervised Semantic Segmentation (ZSWSSS) without using any real images during training. SynthSeg-Agents is built on the synergy between Large Language Models (LLMs), Vision-Language Models (VLMs), and CLIP-based evaluation mechanisms, and consists of two key modules: a \textit{Self-Refine Prompt Agent} and an \textit{Image Generation Agent}.

The Self-Refine Prompt Agent generates diverse and semantically meaningful prompts describing potential segmentation scenes. To ensure quality and prompt-space coverage, it uses iterative refinement, memory-driven diversity enhancement, and CLIP-based semantic filtering. These prompts are then passed to the \textit{Image Generation Agent}, which synthesizes candidate images via VLMs and uses a frozen CLIP model to assess visual fidelity and semantic alignment. High-confidence samples are retained, and a ViT-based classifier is trained on them to relabel the synthetic dataset, generating consistent pseudo-labels for downstream segmentation training.

By decoupling from real image supervision and leveraging LLM-driven generation, SynthSeg-Agents offers a scalable and generalizable paradigm for generating segmentation training data entirely from scratch. Our framework is modular, model-agnostic, and compatible with existing WSSS pipelines, paving the way toward low-cost, zero-supervision segmentation.

Our key contributions are as follows:
\begin{itemize}
    \item We propose a new problem setting—Zero-Shot Weakly Supervised Semantic Segmentation (ZSWSSS)—where segmentation models are trained without using any real images.
    \item We design a Self-Refine Prompt Agent that generates high-quality, diverse prompts through iterative refinement and CLIP-guided semantic filtering, improving the semantic expressiveness of image descriptions.
    \item We develop an Image Generation Agent that leverages vision-language models to synthesize candidate images, employs a frozen CLIP model to assess image–prompt alignment, and integrates a ViT-based classifier to relabel the entire dataset with pseudo-labels.
    \item By integrating both agents into a unified SynthSeg-Agents framework, we demonstrate that purely synthetic training data can enable competitive segmentation performance on both PASCAL VOC 2012 and MS COCO 2014 benchmarks—without using any real-world training images.
\end{itemize}

\begin{figure*}[t] 
\begin{center}
   \includegraphics[width=1\linewidth]{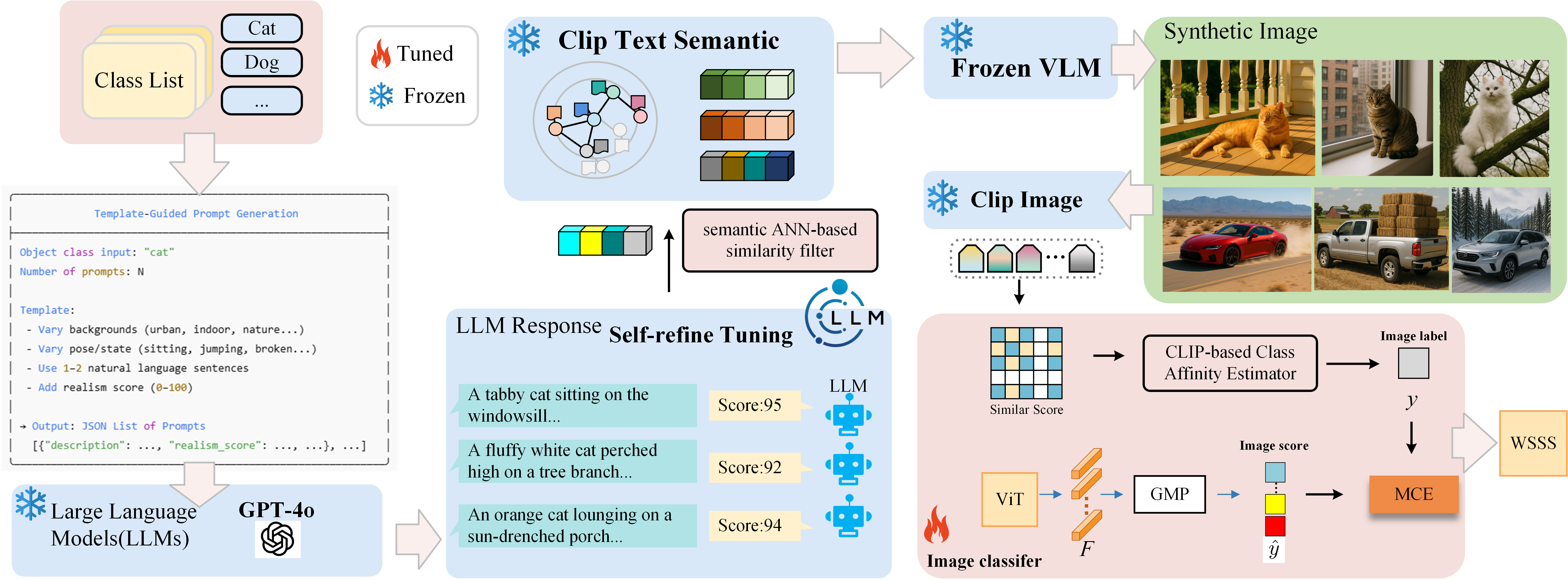}
 % \vspace{0.4cm}
   \caption{The overall pipeline of SynthSeg-Agents. Given a set of target class labels (e.g., {cat, dog, ...}), the Self-Refine Prompt Agent first generates image descriptions using a template-guided prompt mechanism. These prompts are iteratively refined via LLMs and filtered using CLIP-based semantic ANN to enhance relevance and diversity. The selected prompts are passed to the Image Agent, which uses a VLM to generate synthetic images. A frozen CLIP model is used to supervise a lightweight classifier for assigning pseudo-labels to the generated images. The final image–label pairs are used to train a WSSS model without using any real images.}
    \label{fig:framework}
\end{center}
 % \vspace{-0.1cm}
\end{figure*}

\section{Related Work}

\subsection{Weakly Supervised Semantic Segmentation}

Weakly Supervised Semantic Segmentation (WSSS) has emerged as a promising direction for semantic scene understanding, particularly due to its potential to drastically reduce annotation costs by relying on image-level labels or other weak signals. Compared to fully supervised methods requiring costly pixel-wise annotations, WSSS frameworks aim to leverage coarse supervision to guide segmentation model training in a scalable manner. Early seminal work such as CAM~\cite{zhou2016learning} laid the foundation for this research line by highlighting class-specific discriminative regions in feature maps derived from classification networks.

Subsequent works significantly improved mask quality by expanding these activation regions. For instance, AffinityNet~\cite{ahn2018learning}, IRNet~\cite{ahn2019weakly}, and SEAM~\cite{wang2020self} introduced mechanisms such as semantic affinity propagation, inter-pixel relation modeling, and self-supervised equivariant consistency, which helped capture more complete object structures beyond coarse activation peaks. These methods still relied heavily on real image content to anchor their pixel-level predictions.

To further enhance supervision signals, several methods proposed label refinement strategies using pseudo-label self-training~\cite{li2021pseudo}, seeded region growing~\cite{huang2018weakly}, and class-consistent co-segmentation~\cite{fan2018cosegmentation}. These approaches iteratively updated pseudo-masks to approximate ground truth labels, yet remained inherently constrained by the initial localization accuracy provided by the CAM-based seeds.

In parallel, new lines of work attempted to overcome spatial fragmentation and semantic drift in pseudo-labels. For instance, cross-image information propagation~\cite{zhang2020causal,lee2021railroad} was introduced to transfer contextual knowledge between similar objects across images. Others adopted global-local feature modeling~\cite{jiang2022l2g} or uncertainty-aware label correction~\cite{zhang2020reliability} to further suppress spurious activations. While these contributions marked notable progress, they all fundamentally depend on the availability and diversity of real training images. Consequently, the learned supervision remains bounded by the information content and bias in the training corpus. Our work challenges this assumption by proposing a zero-supervision alternative that eliminates the need for any real image input during training.

\subsection{Generative Models for Semantic Segmentation}

Generative models have increasingly been adopted to enhance data diversity or act as auxiliary supervision sources for segmentation tasks. GAN-based frameworks~\cite{souly2017semi, hung2018adversarial} introduced adversarial learning to either generate synthetic images or regularize segmentation masks, proving especially useful in semi-supervised settings. These models aimed to produce realistic samples conditioned on semantic maps or latent codes, but still required initial training on real data.

Recently, diffusion models have gained traction due to their ability to generate high-fidelity, diverse images through iterative denoising processes. Models such as DiffusionSeg~\cite{li2023diffusionseg} and MaskDiffusion~\cite{xu2023masked} explored these techniques in dense prediction contexts, where synthetic images or masks were used to supplement the training set. Despite their impressive image generation quality, these methods are either trained jointly with segmentation models or require real samples to anchor the generative process.

In the context of weak supervision, the advent of open-vocabulary segmentation frameworks~\cite{xu2023open,zhang2023k} leveraged pretrained vision-language models (VLMs), such as CLIP~\cite{radford2021learning} and BLIP~\cite{li2022blip}, to extend segmentation capabilities to unseen classes. This is typically achieved by mapping text embeddings to visual representations or through prompt-based conditioning~\cite{lu2023openclipseg,rao2022denseclip}. However, these methods still rely on real visual inputs and do not address training from scratch using purely synthetic data.

While synthetic data generation has found success in image classification~\cite{azizi2023synthetic} and object detection~\cite{fang2023generate}, dense prediction tasks such as semantic segmentation present unique challenges due to the requirement for fine-grained pixel-level correspondence. To the best of our knowledge, no prior work fully eliminates real images in the WSSS setting. Our approach is among the first to establish a zero-shot segmentation pipeline powered by large-scale generative capabilities of LLMs and VLMs.

\subsection{Large Language Models in Vision}

Large Language Models (LLMs) have recently revolutionized multiple fields through their strong capabilities in reasoning, abstraction, and text-based generation. In the vision domain, their use has been extended to tasks such as image captioning~\cite{yang2022zerocap}, instruction generation~\cite{yang2023gpt4tools}, and synthetic QA data creation for document understanding~\cite{chen2023synthdoc}. Models like GPT-4~\cite{openai2023gpt4} have demonstrated multi-modal reasoning through language grounding, while systems such as Visual ChatGPT~\cite{liu2023visualchatgpt} integrate LLMs with visual modules for interactive decision-making.

A growing line of research also explores prompt engineering as a means to adapt pretrained vision-language models to new downstream tasks without fine-tuning. Approaches such as CoOp~\cite{zhou2022learning} and PromptDet~\cite{feng2023promptdet} learn input prompts that guide inference, improving flexibility and generalization in open-world settings. Still, these methods do not exploit LLMs for structured training data creation.

Recent tool-augmented frameworks like HuggingGPT~\cite{shen2023hugginggpt} chain together multiple specialized models using LLMs as planners or coordinators. 
However, these agent-style systems mostly focus on high-level reasoning or retrieval. 
In contrast, our LLM-powered multi-agent framework generates structured dense-prediction traoining data. 
By combining prompt refinement, VLM-based synthesis, and CLIP-based selection, we train segmentation models without a single real image.

% In contrast, our method leverages the generative capabilities of LLMs in a multi-agent framework to create structured training data for dense prediction. 
% By integrating prompt refinement, VLM-based synthesis, and CLIP-guided selection, we show that segmentation models can be trained without seeing a single real image.

\section{Method}
\label{sec:method}

\subsection{Overview}

As illustrated in Fig.~\ref{fig:framework}, we propose ~\textit{SynthSeg-Agents}, a modular multi-agent framework for Zero-Shot Weakly Supervised Semantic Segmentation (ZSWSSS). 
The pipeline comprises two core components: Self-Refine Prompt Agent and  Image Generation Agent. 
% The former generates diverse and semantically rich image descriptions using a template-guided mechanism and an LLM-based iterative refinement process. 
% To ensure quality and diversity, the generated prompts are filtered through a CLIP-based semantic ANN module.

The filtered prompts are then passed to the Image Agent, which leverages a vision-language model (VLM) to synthesize images. A frozen CLIP model is used to evaluate semantic consistency between prompts and images. Only high-confidence samples are retained, and a ViT-based classifier is trained on these samples to propagate labels across the entire synthetic dataset. The final output is a fully labeled synthetic image set, which is used to train a segmentation model from scratch without any real images.

The design of SynthSeg-Agents follows a modular and decoupled architecture, where each component operates independently yet contributes toward the shared goal of generating high-quality training data. This modularity allows for flexible substitution and future extension. For instance, the prompt generation module can be upgraded with more capable LLMs, or the VLM image generator can be replaced based on domain specificity, making the framework adaptable to both generic and task-specific segmentation pipelines.

\subsection{Self-Refine Prompt Agent}
\label{sec:Prompt}

Given a target object class $c \in \mathcal{C}$ (e.g., ``cat''), we design a Self-Refine Prompt Agent to generate a diverse and semantically meaningful set of image descriptions. This process begins with a structured prompt template function $\mathcal{T}(\cdot)$, which encodes constraints on background variation, pose/state diversity, natural language fluency, and realism scoring. An LLM instantiates the template with class $c$ to produce an initial prompt set:
\begin{equation}
P_{\text{init}} = \mathcal{T}(c).
\end{equation}

To increase prompt diversity and prevent redundancy, we introduce an iterative self-refinement loop. At each iteration, the LLM generates new candidates by expanding or refining previous prompts. Specifically, we refine low-quality prompts by prompting the LLM to rephrase them with richer backgrounds, varied object states, or added context—enhancing diversity while preserving semantics. This process is guided by a memory buffer $\mathcal{M}$ that stores both the textual content and CLIP text embeddings of all previously accepted prompts. For each new prompt $p$, we compute its CLIP embedding $f_{\text{clip}}(p)$ and retrieve its nearest neighbor from $\mathcal{M}$ using approximate nearest neighbor (ANN) search (e.g., FAISS). The cosine similarity to its nearest neighbor is computed as:
\begin{equation}
\text{score}(p) = \cos\left( f_{\text{clip}}(p), f_{\text{clip}}(p') \right), \quad p' \in \mathcal{M}.
\end{equation}
Only prompts with similarity score below a threshold $\delta$ are considered sufficiently novel and added to the final prompt set and memory buffer:
\begin{equation}
P_{\text{refined}} = \{ p \in P_{\text{init}} \mid \text{score}(p) < \delta \}.
\end{equation}

This memory-based filtering mechanism ensures that prompt generation avoids semantic collapse and maintains a diverse distribution, improving the utility of generated images while minimizing wasteful samples.

The full generation and refinement process is described in Algorithm~\ref{alg:SR}. This self-refinement and filtering process not only reduces redundancy but also maximizes semantic coverage across object poses, backgrounds, and contexts, ensuring downstream image generation benefits from high prompt diversity.

\begin{algorithm}[t]
\caption{{\small{Self-Refine for prompt generation with CLIP filtering}}}
\label{alg:SR}
\SetKwInOut{Input}{Input}
\SetKwInOut{Output}{Output}

\Input{Category label $c$, model GPT-4o, prompt templates $\{p_{\text{gen}}, p_{\text{refine}}\}$, quality threshold $\epsilon$, diversity threshold $\delta$}
\Output{Filtered prompt set $P_{\text{refined}}$}

$y_0 \leftarrow \text{GPT-4o}(\text{Initial\_prompt}(p_{\text{gen}}, c))$\;
\Repeat{$\text{score}_{y_0} < \epsilon$}{
    $\text{score}_{y_0} \leftarrow \text{GPT-4o}(\text{Refine\_prompt}(p_{\text{refine}}, c, y_0))$\;
    $y_0 \leftarrow \text{GPT-4o}(\text{Initial\_prompt}(p_{\text{gen}}, c))$\;
}
$P \leftarrow \{y_0\}$\;

$P_{\text{refined}} \leftarrow \{\}$, $\mathcal{M} \leftarrow \{\}$\;
\ForEach{$p \in P$}{
    $e_p \leftarrow f_{\text{clip}}(p)$\;
    $e_{nn} \leftarrow \text{ANN}(e_p, \mathcal{M})$\;
    $\text{score} \leftarrow \cos(e_p, e_{nn})$\;
    \If{$\text{score} < \delta$}{
        $P_{\text{refined}} \leftarrow P_{\text{refined}} \cup \{p\}$\;
        $\mathcal{M} \leftarrow \mathcal{M} \cup \{p\}$\;
    }
}
\Return $P_{\text{refined}}$\;
\end{algorithm}

\begin{figure}[t]
\centering
\includegraphics[width=0.9\linewidth]{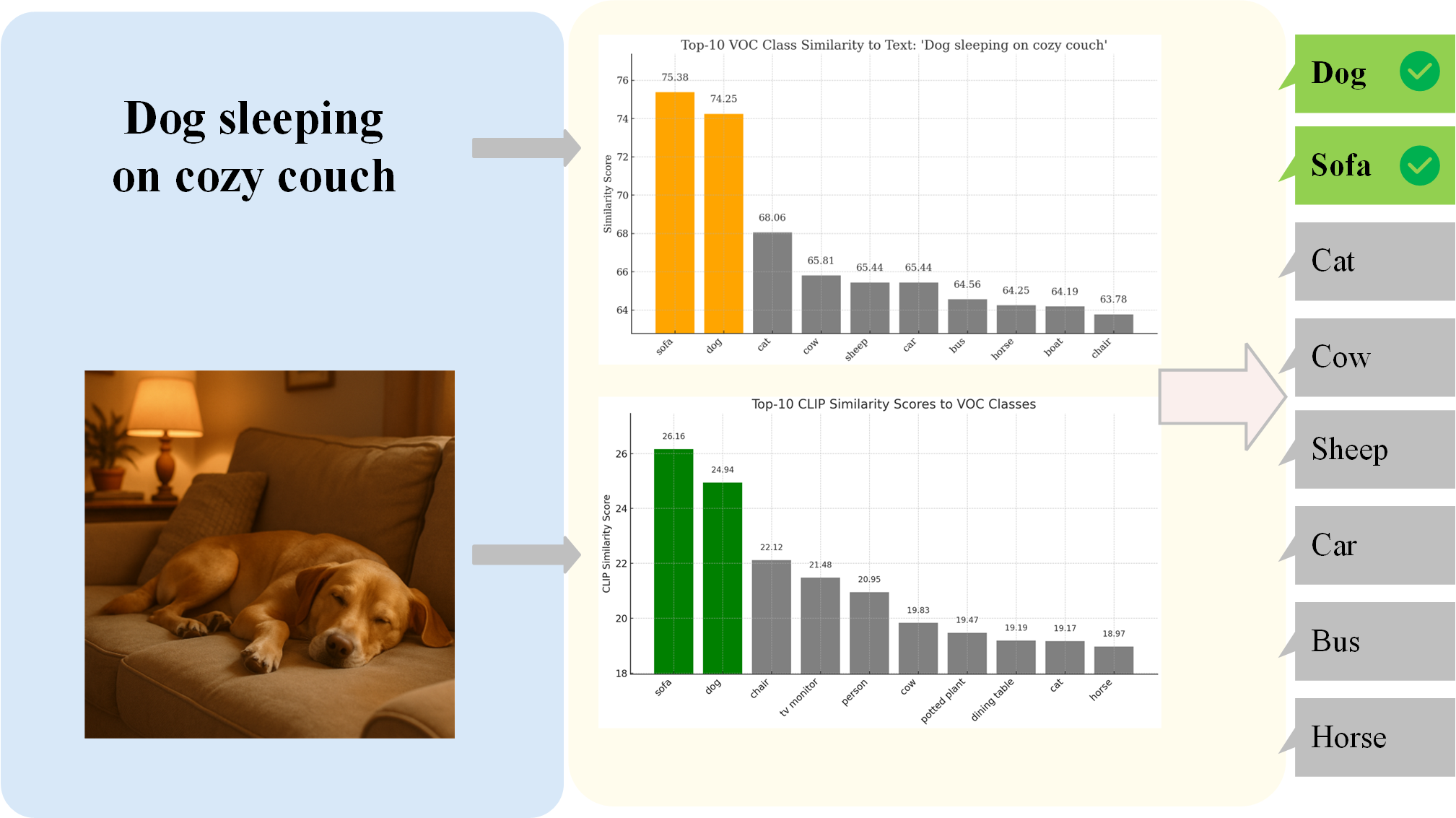}
% \vspace{-0.3cm} 
\caption{
\textbf{CLIP-based pseudo-labeling.} 
Given a generated image and its prompt, we compute CLIP similarity between the prompt/image and all VOC classes. Classes with high scores in both spaces are selected as pseudo-labels—in this case, \textbf{dog} and \textbf{sofa}.
}

\label{fig:clip_dual_alignment}
\vspace{-0.15cm} 
\end{figure}

\subsection{Image Generation Agent}
\label{sec:ImageAgent}

Following the prompt generation stage, the next key component in the SynthSeg-Agents framework is the Image Agent. This agent is responsible for converting textual descriptions into visual data by synthesizing images and assigning semantic labels, making them suitable for downstream segmentation training. It serves as the critical bridge that transforms language-driven prompts into structured image–label pairs.

\vspace{0.5em}
\paragraph{Image synthesis and CLIP-based filtering.} 
Given a refined prompt set $P_{\text{refined}} = \{p_1, p_2, \dots, p_N\}$, the Image Agent uses a pretrained Vision-Language Model (VLM) $f_{\text{VLM}}(\cdot)$ to generate synthetic images from textual prompts. For each prompt, an image is synthesized as follows:
\begin{equation}
I_{\text{gen}} = f_{\text{VLM}}(p_i), \quad \forall p_i \in P_{\text{refined}}.
\end{equation}
This generation process yields a diverse set of visual samples that are intended to reflect the semantics embedded in their textual descriptions.

To assess alignment between the prompt and all class labels, we first compute CLIP-based text–text similarity. The prompt $p_i$ and all class names $\texttt{label}(c)$ for $c \in \mathcal{C}$ are embedded using a frozen CLIP text encoder:
\begin{equation}
\mathbf{e}_T = f_{\text{clip}}(p_i), \quad \mathbf{e}_c = f_{\text{clip}}(\texttt{label}(c)).
\end{equation}
Each embedding is $\ell_2$-normalized, and the similarity is calculated as:
\begin{equation}
\text{score}_{\text{text}}(p_i, c) = \frac{1 + \tilde{\mathbf{e}}_T^\top \tilde{\mathbf{e}}_c}{2}.
\end{equation}

We select all candidate labels with text similarity greater than a threshold $\gamma_1$:
\begin{equation}
\mathcal{C}_{\text{text}} = \left\{ c \in \mathcal{C} \;\middle|\; \text{score}_{\text{text}}(p_i, c) > \gamma_1 \right\}.
\end{equation}

We then perform a second validation to confirm whether each candidate label is also semantically aligned with the generated image. The image is encoded using CLIP:
\begin{equation}
\mathbf{e}_I = f_{\text{clip}}(I_{\text{gen}}), \quad \tilde{\mathbf{e}}_I = \frac{\mathbf{e}_I}{\|\mathbf{e}_I\|_2}.
\end{equation}
The similarity between image and candidate label is:
\begin{equation}
\text{score}_{\text{image}}(I_{\text{gen}}, c) = \frac{1 + \tilde{\mathbf{e}}_I^\top \tilde{\mathbf{e}}_c}{2}.
\end{equation}
To visualize this dual-filtering mechanism, we present an example in Fig.~\ref{fig:clip_dual_alignment}. Given the prompt ``Dog Sleeping on coze couch'' and its generated image, we compute CLIP similarities between both the prompt and the image against all VOC classes. Classes with high scores in both spaces—such as \texttt{dog} and \texttt{sofa}—are retained as pseudo-labels.

All candidate labels from $\mathcal{C}_{\text{text}}$ are sorted by their $\text{score}_{\text{image}}$, and the top-$N$ scoring labels are retained:
\begin{equation}
\mathcal{C}_{\text{high}} = \text{Top}_N \left\{ c \in \mathcal{C}_{\text{text}} \;\middle|\; \text{score}_{\text{image}}(I_{\text{gen}}, c) \right\}.
\end{equation}
Each $(I_{\text{gen}}, c)$ pair where $c \in \mathcal{C}_{\text{high}}$ is accepted as a high-quality pseudo-labeled training sample:
\begin{equation}
\mathcal{D}_{\text{high}} = \left\{(I_{\text{gen}}, c) \mid c \in \mathcal{C}_{\text{high}} \right\}.
\end{equation}

\begin{algorithm}[t]
\caption{{\small{Image Agent: image synthesis, filtering, and relabeling}}}
\label{alg:ImageAgent}
\SetKwInOut{Input}{Input}
\SetKwInOut{Output}{Output}

\Input{Refined prompts $P_{\text{refined}}$, VLM $f_{\text{VLM}}$, CLIP model $f_{\text{clip}}$, thresholds $\gamma_1$, $N$}
\Output{Relabeled image–label pairs $\{(I_{\text{gen}}, L_{\text{gen}})\}$}

\ForEach{$p_i \in P_{\text{refined}}$}{
    $I_{\text{gen}} \leftarrow f_{\text{VLM}}(p_i)$\;
    \ForEach{$c \in \mathcal{C}$}{
        $s_1 \leftarrow \text{score}_{\text{text}}(p_i, c)$\;
        \If{$s_1 > \gamma_1$}{
            $s_2 \leftarrow \text{score}_{\text{image}}(I_{\text{gen}}, c)$\;
            \If{$c \in \text{Top}_N(s_2)$}{
                $\mathcal{D}_{\text{high}} \leftarrow \mathcal{D}_{\text{high}} \cup \{(I_{\text{gen}}, c)\}$\;
            }
        }
    }
}
Train ViT classifier on $\mathcal{D}_{\text{high}}$\;
\ForEach{$I_{\text{gen}}$}{
    $L_{\text{gen}} \leftarrow \text{Classifier}(I_{\text{gen}})$\;
}
\Return $\{(I_{\text{gen}}, L_{\text{gen}})\}$\;
\end{algorithm}

\vspace{0.5em}
\paragraph{Training a classifier for relabeling.} 
After high-confidence samples have been selected, we use them to train a patch-wise ViT-based image classifier. This classifier serves to relabel the entire synthetic dataset with consistent and scalable supervision. Specifically, each image $X_{\text{in}}$ is divided into $s = \frac{hw}{d^2}$ non-overlapping patches $X_{\text{patch}} \in \mathbb{R}^{d \times d \times 3}$ and embedded using a Vision Transformer to obtain:
\begin{equation}
F = \text{ViT}(X_{\text{in}}) \in \mathbb{R}^{s \times e}
\end{equation}
where $F$ denotes the patch-level embeddings and $e$ is the embedding dimension.

We then apply a linear classifier $W \in \mathbb{R}^{e \times |\mathcal{C}|}$ to obtain class logits:
\begin{equation}
Z = \text{softmax}(F W).
\end{equation}
These logits are aggregated across patches using global max pooling to yield image-level predictions $\hat{y} \in \mathbb{R}^{1 \times |\mathcal{C}|}$. The classifier is trained using a multi-label binary cross-entropy loss:
\begin{equation}
\mathcal{L}_{\text{MCE}} = -\frac{1}{|\mathcal{C}|} \sum_{c \in \mathcal{C}} \left[ y_c \log(\hat{y}_c) + (1 - y_c) \log(1 - \hat{y}_c) \right].
\end{equation}
Once training is complete, the classifier is used to relabel all synthesized images, including those outside the high-confidence set. The resulting relabeled dataset $\{(I_{\text{gen}}, L_{\text{gen}})\}$ serves as input to downstream segmentation models. The complete image generation and labeling procedure is summarized in Algorithm~\ref{alg:ImageAgent}.

\subsection{Synthetic Dataset for ZSWSSS}
\label{sec:TrainingZSWSSS}

Let $\mathcal{D}_{\text{ZSWSSS}} = \{(I_i, L_i)\}_{i=1}^{N}$ denote the synthetic training dataset produced by SynthSeg-Agents, where $I_i \in \mathbb{R}^{H \times W \times 3}$ is a synthesized image and $L_i \in \mathbb{R}^{|\mathcal{C}|}$ is its corresponding image-level pseudo-label over class set $\mathcal{C}$. These labels are obtained via the image relabeling module described in Section~\ref{sec:ImageAgent}.

\begin{figure}[t]
\centering
\includegraphics[width=1.0\linewidth]{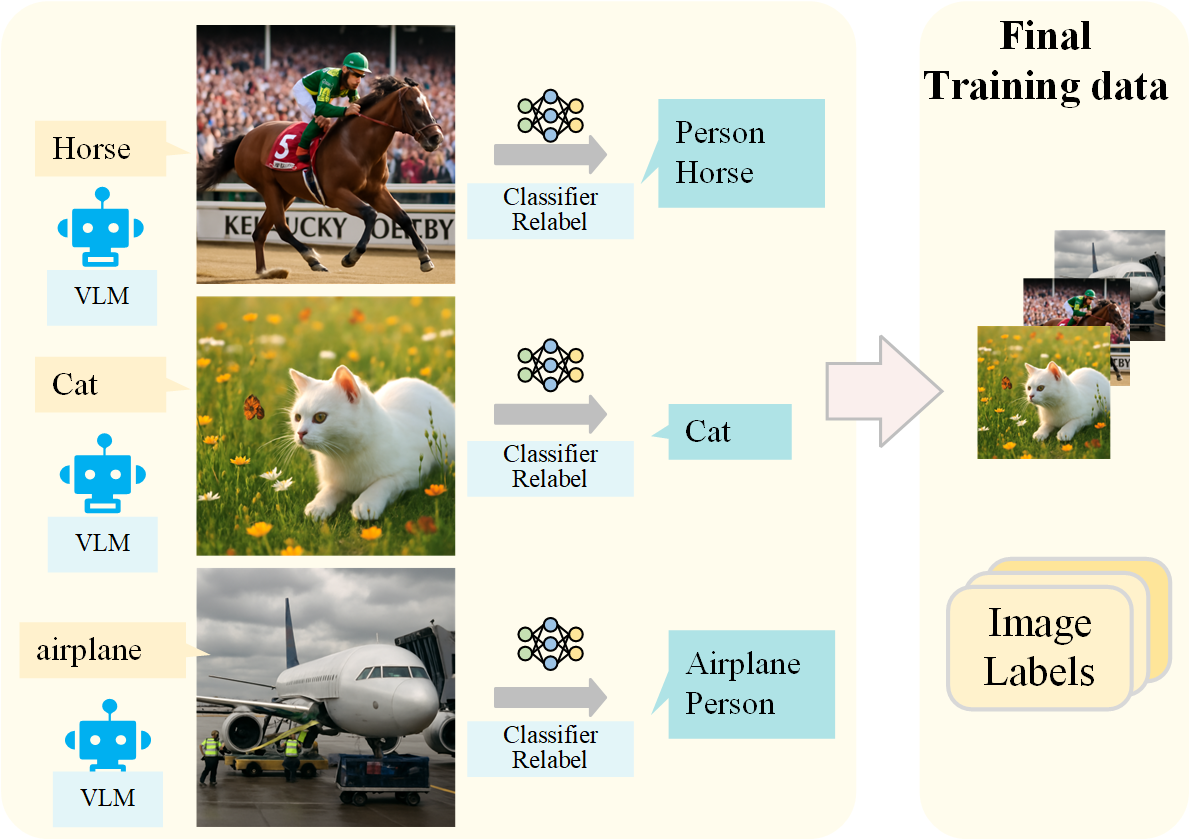}
% \vspace{-0.3cm} 
\caption{The final training data consists of synthetic images generated by VLMs, followed by relabeling using our image classifier. For example, when prompting for the horse class, the VLMs may generate images that include multiple objects. To address this, we apply our image classifier to relabel each image with a refined to ensure that the final dataset is cleaner and more suitable for downstream training.}

\label{fig:traindata}
\vspace{-0.3cm} 
\end{figure}

As illustrated in Fig.~\ref{fig:traindata}, this dataset conforms to the standard input format expected by existing Weakly Supervised Semantic Segmentation (WSSS) pipelines:
\begin{equation}
\mathcal{D}_{\text{ZSWSSS}} \approx \mathcal{D}_{\text{WSSS}}^{\text{real}} = \{(I_j^{\text{real}}, L_j^{\text{image}})\}
\end{equation}
where $\mathcal{D}_{\text{WSSS}}^{\text{real}}$ typically consists of real images paired with image-level human labels. In our case, both components are synthesized, making the entire training pipeline fully zero-shot and free of any real-world supervision.

Importantly, our method does not propose a new segmentation training algorithm. Instead, we generate synthetic training data that can be seamlessly integrated into any existing WSSS framework—such as SEAM~\cite{wang2020self}, AffinityNet~\cite{ahn2018learning}, or IRNet~\cite{ahn2019weakly}—without modification.
\begin{figure*}[t]
\centering
\includegraphics[width=\linewidth]{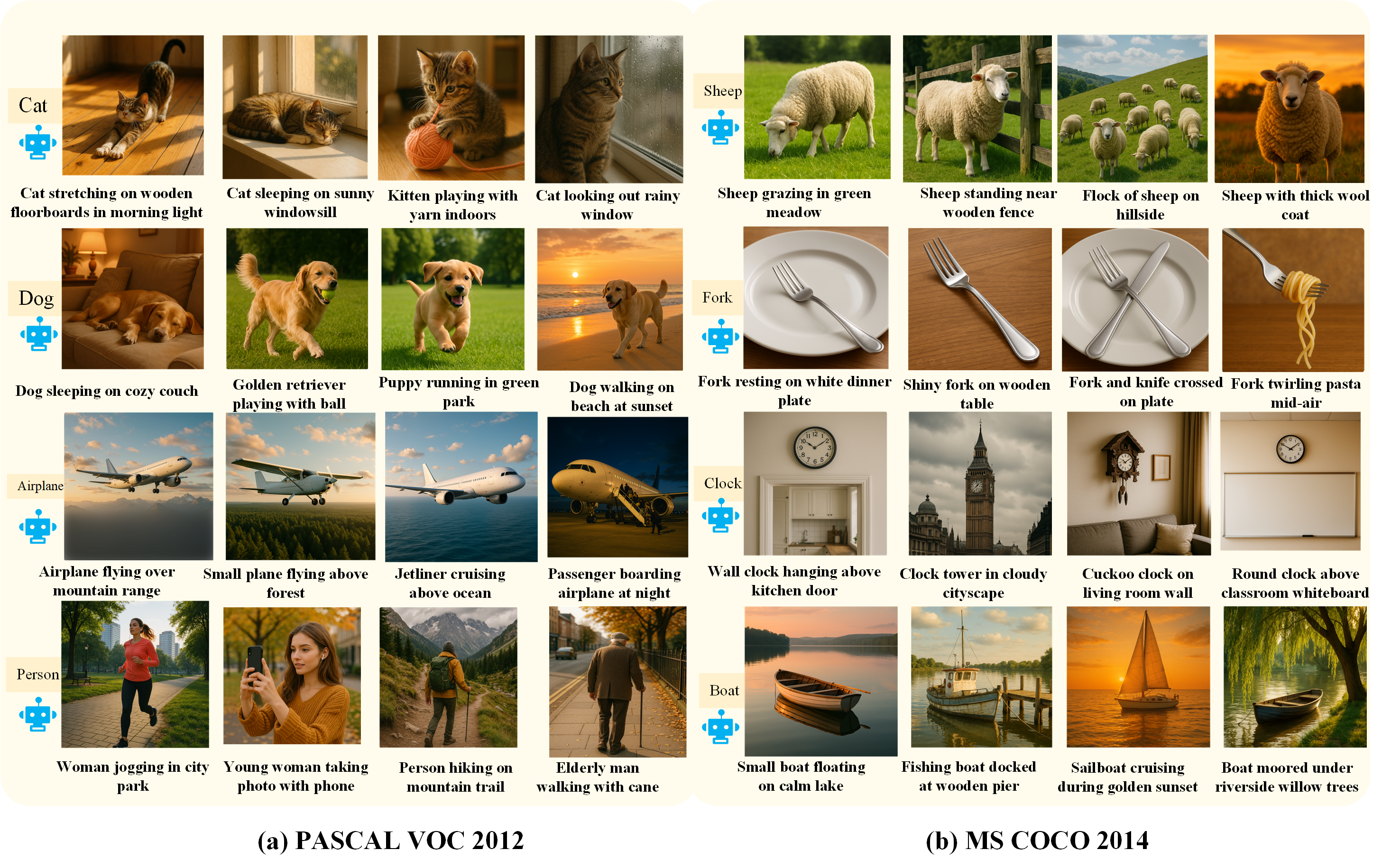}

\caption{
\textbf{Synthetic images generated by SynthSeg-Agents for PASCAL VOC 2012 and MS COCO 2014.} 
\textbf{(a)} Samples for selected PASCAL VOC classes. 
\textbf{(b)} Samples for selected MS COCO classes. 
Each pair shows the class label (left) and the corresponding synthetic image (right), generated using prompts from our Self-Refine Prompt Agent.
}

\label{fig:dfcase}
\vspace{0.4cm} 
\end{figure*}
To train a segmentation model $\mathcal{M}_{\text{seg}}$, we use standard WSSS pipelines:
\begin{equation}
\mathcal{M}_{\text{seg}} = \text{Train}(\mathcal{D}_{\text{ZSWSSS}})
\end{equation}
where $\mathcal{M}_{\text{seg}}$ is any model (e.g., DeepLab or SEAM) trained from scratch on $\mathcal{D}_{\text{ZSWSSS}}$.

This step produces synthetic data compatible with standard WSSS methods—no real images needed—making it a scalable, drop-in zero-shot generator of weakly labeled segmentation data.

% This final step completes the SynthSeg-Agents pipeline, demonstrating that our synthetic data is compatible with established WSSS methods while eliminating the need for real images. Our framework is thus a drop-in, scalable generator of weakly labeled segmentation data in the zero-shot regime.

\section{Experiments}
\label{sec:experiments}

In this section, we evaluate the effectiveness of the proposed \textit{SynthSeg-Agents} framework for Zero-Shot Weakly Supervised Semantic Segmentation (ZSWSSS). Our experiments aim to answer the following key questions:

\begin{itemize}
    \item Can synthetic training data generated without any real images yield competitive segmentation results? (Sec.~\ref{sec:finalresults})
    \item Can synthetic images serve as effective auxiliary training data to improve the performance of existing real-image-based WSSS methods? (Sec.~\ref{sec:finalresults})
    \item What is the impact of each component in the proposed multi-agent pipeline? (Sec.~\ref{sec:ablation})
\end{itemize}

\begin{figure*}[t]
\centering
\includegraphics[width=1.0\linewidth]{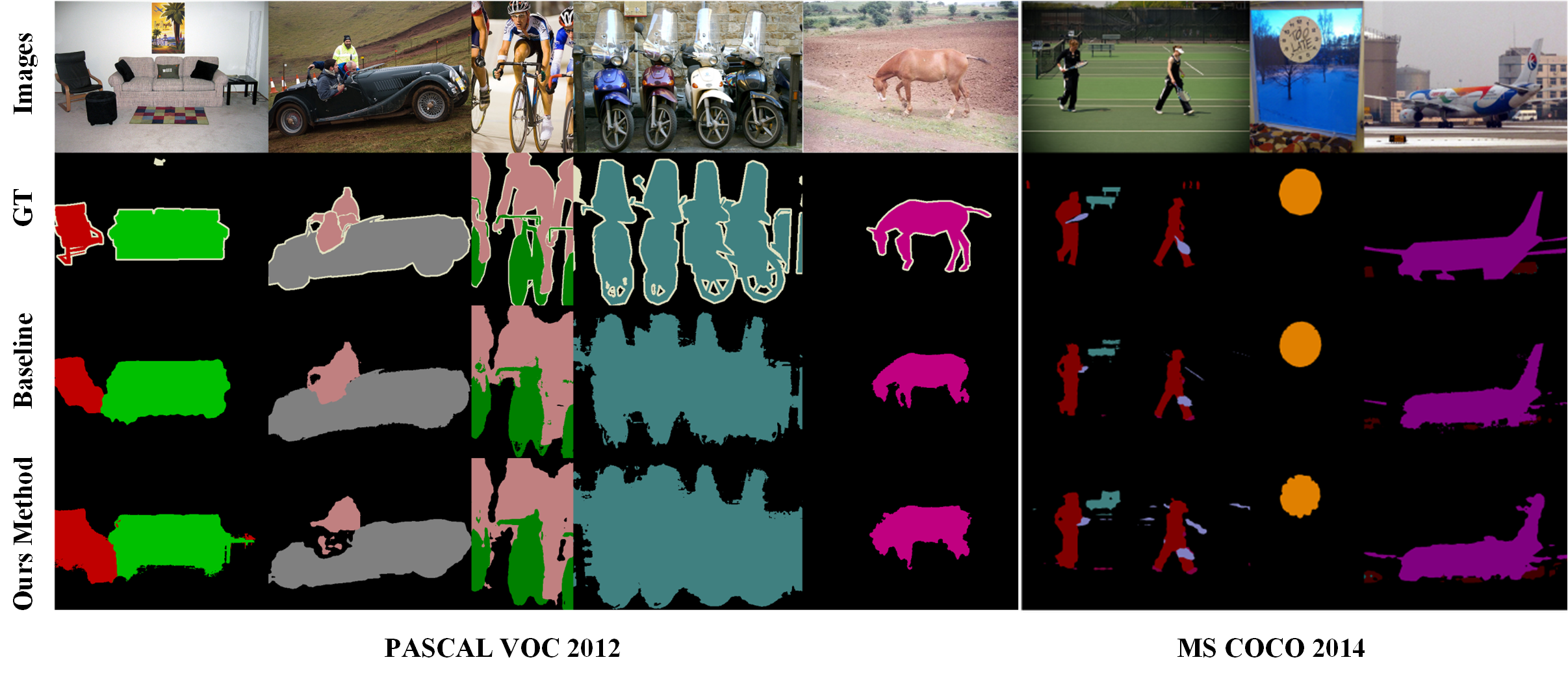} 
\caption{\textbf{Qualitative comparison of segmentation results on PASCAL VOC 2012 and MS COCO 2014.} From top to bottom: input image, ground truth (GT), baseline method ToCo~\cite{ru2023token}, and our SynthSeg-Agents results. 
While ToCo is trained with real images and pixel-level supervision, our method uses only synthetic data generated via LLM/VLM agents without any real images. Despite this, SynthSeg-Agents produces semantically meaningful and spatially coherent segmentation maps, demonstrating the potential of fully synthetic training for zero-shot WSSS.
}
\label{fig:result}
\vspace{0.1cm} 
\end{figure*}
\subsection{Datasets and Evaluation}
Following previous works~\cite{ru2023token,Peng_2023_ICCV} in weakly supervised semantic segmentation, we evaluate our method on the PASCAL VOC 2012~\cite{everingham2010pascal} and MS COCO 2014~\cite{lin2014microsoft} benchmarks. Unlike prior approaches, we train our models entirely on synthetic data generated by SynthSeg-Agents and evaluate on the standard real validation sets.

For PASCAL VOC 2012, we generate 10k synthetic images covering 20 foreground categories. For MS COCO 2014, we synthesize 80k training images spanning all 80 object classes. No real images are used during training.

All segmentation models are trained from scratch using the generated image--label pairs. We report performance using the standard mean intersection-over-union (\textit{mIoU}) metric~\cite{cheng2022masked}.

\subsection{Implementation Details}
\label{sec:implementation}

We implement the Self-Refine Prompt Agent using GPT-4o and generate prompts for each object class based on structured class-specific templates. The prompt quality acceptance threshold is set to $\epsilon = 0.95$, and the semantic diversity filtering threshold is set to $\delta = 0.92$ to ensure sufficient prompt variation. For image generation, we employ GPT-Image-1 as the Vision-Language Model (VLM), and use CLIP (ViT-B/32) to compute image--text alignment scores. The cosine similarity score is scaled to $[0, 1]$, and a threshold of $\gamma = 0.7$ is used to retain only high-confidence image--prompt pairs.

To relabel the synthetic dataset, we train a ViT-B/32-based image classifier on the high-confidence subset $\mathcal{D}_{\text{high}}$. Each input image is resized to $384 \times 384$ and partitioned into $24 \times 24$ patches, as suggested in SEED~\cite{kolesnikov2016seed}, for input to the ViT encoder. During inference, images are resized to $960 \times 960$ for improved spatial granularity. The classifier is trained using a batch size of 16 for up to 50 epochs on two NVIDIA RTX 4090 GPUs. We use the Adam optimizer, with an initial learning rate of $10^{-3}$ for the first two epochs and $10^{-4}$ for the remaining training period. For downstream WSSS tasks, we evaluate the segmentation performance of our synthetic dataset using several widely adopted frameworks.

\begin{table}[t]
\caption{\textbf{Zero-Shot WSSS Performance on PASCAL VOC.} 
\textit{Seen} / \textit{Unseen} denote mIoU for seen and unseen classes. 
"Real" uses only real images; "Synth" uses only synthetic data. }

\centering
\small % 字体介于 scriptsize 和 normalsize 之间
\setlength{\tabcolsep}{5pt} % 比默认 6pt 稍小一点，压缩列间距
\renewcommand{\arraystretch}{1.25} % 稍微加点行距，保持可读性
\begin{tabular}{l|cc|cc}
    & \multicolumn{2}{c|}{Train Set/\%} & \multicolumn{2}{c}{mIoU/\%} \\
    Methods & Type & Categories & Seen & Unseen \\

    \midrule
    \multicolumn{5}{l}{\textit{\textbf{Weakly Supervised Semantic Segmentation} }}\\
    SAS~\cite{kim2023semantic}  & Real & 20 & 70.2 & -- \\
    FPR~\cite{chen2023fpr}  & Real & 20 & 70.2 & -- \\
    ToCo~\cite{ru2023token}  & Real & 20 & 70.2 & -- \\
    IACD~\cite{wu2023image} & Real & 20 & 71.4 \\
    Seco~\cite{Yang_2024_CVPR}  & Real & 20 & 74.0 & -- \\
    SFC~\cite{zhao2024sfc}  & real & 20 & 71.2 & -- \\
    GPCD~\cite{wu2025generative} & real & 20 & 72.3 & -- \\
    APC~\cite{wu2025adaptive} & real & 20 & 72.3 & -- \\
    SynthSeg-Agents (Seco) & Synth + Real & 20 & \textbf{75.4}& --  \\
    \midrule
    \multicolumn{5}{l}{\textit{\textbf{\underline{Synthetic Image (Zero-shot)}}}} \\
    SynthSeg-Agents (Toco) & Synth & 20 & -- & 57.4 \\
    SynthSeg-Agents (Seco) & Synth & 20 & -- & \textbf{60.1} \\

\end{tabular}

\label{tab:voc_results}
\end{table}

\subsection{Final Segmentation Results}
\label{sec:finalresults}

We evaluate the segmentation performance of \textit{SynthSeg-Agents} on two widely adopted benchmarks: PASCAL VOC 2012 and MS COCO 2014. Unlike traditional WSSS methods, our framework is trained entirely on synthetic image--label pairs without using any real image supervision.

\paragraph{PASCAL VOC 2012.} 
As shown in Tab.~\ref{tab:voc_results}, we compare our method against fully supervised and weakly supervised approaches. While SIGN~\cite{cheng2021sign} achieves 41.3\% mIoU on unseen classes with full supervision, our purely synthetic framework achieves 57.4\% and 60.1\% mIoU when integrated into the ToCo~\cite{ru2023token} and Seco~\cite{Yang_2024_CVPR} pipelines, respectively—without using any real training images. This demonstrates the effectiveness of our multi-agent synthetic data generation approach in the ZSWSSS setting. Additionally, finetuning with real images further boosts performance to 75.4\% on seen classes, illustrating the complementary value of our synthetic dataset, though this is not the main focus of our work.

\paragraph{MS COCO 2014.} 
On COCO, we further validate the generalization of our synthetic dataset. As shown in Tab.~\ref{tab:coco_results}, our method achieves 30.2\% mIoU when trained purely with synthetic data. After finetuning with real COCO images, the performance improves to 47.8\%, surpassing state-of-the-art WSSS methods such as ToCo and Seco. This indicates that our synthetic images serve as strong pretraining data even when integrated with real image supervision.

\paragraph{Qualitative Results.} 
We visualize example segmentation results in Fig.~\ref{fig:result}. Compared to the baseline ToCo model trained with real images, SynthSeg-Agents produces semantically coherent masks despite using only synthetic training data. In addition, Fig.~\ref{fig:dfcase} showcases synthetic images generated for various PASCAL VOC and COCO classes, illustrating the diversity and semantic alignment of our synthesized training set.

These results demonstrate that SynthSeg-Agents enables competitive and scalable WSSS purely through LLM- and VLM-driven synthetic data generation.

\begin{table}[t]
\caption{\textbf{Semantic segmentation performance on the COCO dataset.} mIoU results with different training data size and model configurations.}
\centering
\small
\setlength{\tabcolsep}{5pt}
\renewcommand{\arraystretch}{1.25}
\begin{tabular}{l|l|c|c|c}
Method & Train Set & Number & Backbone & mIoU \\
\midrule
\multicolumn{5}{l}{\textit{\textbf{Train with Pure Real Data (WSSS)}}} \\
SIPE ~\cite{chen2022self} & COCO & 80k (all) & ViT-B & 43.6 \\
TSCD ~\cite{Xu_Wang_Sun_Xu_Meng_Zhang_2023} & COCO & 80k (all) & MiT-B1  & 40.1 \\
ToCo~\cite{ru2023token} & COCO & 80k (all) & ViT-B & 42.3 \\
SFC~\cite{zhao2024sfc} & COCO & 80k (all) & ViT-B & 44.6 \\
Seco~\cite{Yang_2024_CVPR} & COCO & 80k (all) & ViT-B & 46.7 \\
IAA~\cite{wu2025image} & COCO & 80k (all) & ViT-B & 45.8 \\
APC~\cite{wu2025adaptive} & COCO & 80k (all) & ViT-B & 45.7 \\
\midrule
\multicolumn{5}{l}{\textit{\textbf{Train with Pure Synthetic Data}}} \\
SynthSeg-Agents & Synthetic & 80k & ViT-B & 30.2 \\
\midrule
\multicolumn{5}{l}{\textit{\textbf{Finetune with Real Data}}} \\
SynthSeg-Agents & Synthetic + COCO & 80k + 80k & ViT-B & \textbf{47.8} \\
\end{tabular}
% \vspace{0.4cm} 
\label{tab:coco_results}
\end{table}

\begin{table}[t]
\caption{\textbf{Ablation study of the Self-Refine Prompt Agent on PASCAL VOC.} We compare prompt generation strategies by incrementally adding self-refinement scoring and CLIP-based filtering.}
\centering
\small
\setlength{\tabcolsep}{5pt}
\renewcommand{\arraystretch}{1.25}
\begin{tabular}{l|c|c|c|c}
\textbf{Experiment} & Template & Refine & CLIP & mIoU \\
\midrule
Template only & \checkmark & \xmark & \xmark & 48.1 \\
+ Self-Refine Score & \checkmark & \checkmark & \xmark & 49.9 \\
+ Self-Refine + CLIP & \checkmark & \checkmark & \checkmark & \textbf{52.5} \\
\end{tabular}
\label{tab:ablation_prompt}
\end{table}

\subsection{Ablation Study}
\label{sec:ablation}

To better understand the contribution of each component in the SynthSeg-Agents framework, we conduct a series of ablation experiments on the PASCAL VOC 2012 dataset using a fixed training budget of 2k synthetic images. We separately examine the impact of the 	extbf{Self-Refine Prompt Agent} and the 	extbf{Image Relabeling Module}.
\vspace{-0.2cm} 
\paragraph{Effectiveness of Self-Refine Prompt Agent.}
Tab.~\ref{tab:ablation_prompt} shows that using only template-based prompts yields 48.1 mIoU. Adding self-refinement improves performance to 49.9, while further applying CLIP-based filtering boosts it to 52.5. This confirms the importance of iterative refinement and diversity filtering.

\vspace{-0.2cm} 
\paragraph{Effectiveness of Image Relabeling.}
Tab.~\ref{tab:ablation_prompt} shows that using only template-based prompts yields 48.1 mIoU. Adding self-refinement improves performance to 49.9, while further applying CLIP-based filtering boosts it to 52.5. This confirms the importance of iterative refinement and diversity filtering.

\vspace{-0.2cm} 
\paragraph{Joint Analysis of Both Modules.}
Tab.~\ref{tab:joint_ablation} illustrates the combined effect of both modules. Enabling both prompt refinement and image relabeling leads to the best result of 52.5 mIoU, validating the synergy of modular yet collaborative agents.

\begin{table}[t]
\caption{\textbf{Ablation study of the Image Agent on PASCAL VOC.} We evaluate how prompt-aligned filtering (CLIP) and image relabeling contribute to pseudo-label quality.}
\centering
\small
\setlength{\tabcolsep}{4pt}
\renewcommand{\arraystretch}{1.25}
\resizebox{\columnwidth}{!}{
\begin{tabular}{l|c|c|c|c}
\textbf{Experiment} & Class Label & CLIP Filter & Relabel Classifier & mIoU \\
\midrule
Class Label Only & \checkmark & \xmark & \xmark & 46.7 \\
+ CLIP Filter & \checkmark & \checkmark & \xmark & 50.8 \\
+ CLIP + Relabel (Ours) & \checkmark & \checkmark & \checkmark & \textbf{52.5} \\
\end{tabular}
}

\label{tab:ablation_image}
\end{table}
\begin{table}[t]
\caption{\textbf{Joint ablation of Prompt Agent and Image Agent.} We progressively enable different components across both modules to analyze their combined effect.}
\centering
\small
\setlength{\tabcolsep}{4pt}
\renewcommand{\arraystretch}{1.25}
\resizebox{\columnwidth}{!}{
\begin{tabular}{l|c|c|c|c|c|c}
\textbf{Experiment} & Refine & CLIP (Prompt) & CLIP (Image) & Relabel & Modules Used & mIoU \\
\midrule
Baseline (Template + Class Label) & \xmark & \xmark & \xmark & \xmark & None & 44.1 \\
+ Prompt Refine Only & \checkmark & \checkmark & \xmark & \xmark & Prompt Only & 46.7 \\
+ Image Relabel Only & \xmark & \xmark & \checkmark & \checkmark & Image Only & 48.1 \\
+ Both (Ours) & \checkmark & \checkmark & \checkmark & \checkmark & Full & \textbf{52.5} \\
\end{tabular}
}
\label{tab:joint_ablation}
\end{table}

\vspace{0.5em}
\noindent These results validate the design of SynthSeg-Agents: both the prompt refinement and image relabeling modules contribute substantially to final performance, even under a low-data training regime.

\section{Conclusion}
\label{sec:conclusion}

We introduce \textit{SynthSeg-Agents}, a modular multi-agent framework for Zero-Shot Weakly Supervised Semantic Segmentation (ZSWSSS). By combining a Self-Refine Prompt Agent with a CLIP-guided filtering mechanism and an Image Agent capable of generating and relabeling synthetic data, our method eliminates the need for real training images. Experiments on PASCAL VOC and MS COCO demonstrate that our synthetic data enables competitive performance across multiple WSSS pipelines, while remaining entirely free of real supervision. Ablation studies further validate the effectiveness of both prompt refinement and relabeling components. Our framework offers a scalable and plug-and-play solution for semantic segmentation in scenarios where real data is scarce or unavailable.

\section{Acknowledgements}
This work was supported by the National Natural Science Foundation of China (No. 62471405, 62331003, 62301451), Suzhou Basic Research Program (SYG202316) and XJTLU REF-22-01-010, XJTLU AI University Research Centre, Jiangsu Province Engineering Research Centre of Data Science and Cognitive Computation at XJTLU and SIP AI innovation platform (YZCXPT2022103).
\bibliography{mybib}

\end{document}